%% file: main.tex
\documentclass[twoside]{article}

\makeatletter
\providecommand*{\input@path}{}
\edef\input@path{{styles/}{../styles/}\input@path}
\makeatother

\usepackage{aistats2020}

\PassOptionsToPackage{table}{xcolor}
\RequirePackage{ragged2e}
\usepackage{AuxCommandsAISTATS}
\input{styles/math_commands.tex}

\usepackage{subfiles}

\newcommand{\FedPerAlg}{\textsc{FedPer}}
\newcommand{\FedAvgAlg}{\textsc{FedAvg}}
\newcommand{\sgd}{\text{SGD}}
\NewDocumentCommand{\cifar}{o}{%
	\IfNoValueTF{#1}
	{\text{CIFAR}}
	{\text{CIFAR-#1}}%
}
\newcommand{\flickr}{\text{FLICKR-AES}}
\NewDocumentCommand{\resnet}{o}{%
	\IfNoValueTF{#1}
	{\text{ResNet}}
	{\text{ResNet-#1}}%
}
\NewDocumentCommand{\mobnet}{o}{%
	\IfNoValueTF{#1}
	{\text{MobileNet}}
	{\text{MobileNet-#1}}%
}

\DeclareGraphicsExtensions{.png}
\graphicspath{{final_images/}{../final_images/}}

\begin{document}
\twocolumn[

\aistatstitle{Federated Learning with Personalization Layers}

\aistatsauthor{ Manoj Ghuhan Arivazhagan \And Vinay~Aggarwal }

\aistatsaddress{ Adobe Research \And  Indian Institute of Technology, Roorkee, India } 

\aistatsauthor{Aaditya~Kumar~Singh \And Sunav~Choudhary }

\aistatsaddress{ Indian Institute of Technology, Kharagpur, India \And Adobe Research} ]
\renewcommand{\bibsubfile}[2]{}

\subfile{sections/abstract}

\subfile{sections/introduction}

\subfile{sections/related-work}

\subfile{sections/algorithm}

\subfile{sections/experiments}

\subfile{sections/conclusions}

\bibliographystyle{apalike}
\bibliography{main}

\onecolumn
\subfile{sections/appendix}

\end{document}

%% file: styles/math_commands.tex
\def\eqref#1{equation~\ref{#1}}

\def\1{\bm{1}}

\def\ry{{\textnormal{y}}}

\def\rvx{{\mathbf{x}}}

\DeclareMathAlphabet{\mathsfit}{\encodingdefault}{\sfdefault}{m}{sl}
\SetMathAlphabet{\mathsfit}{bold}{\encodingdefault}{\sfdefault}{bx}{n}
\newcommand{\tens}[1]{\bm{\mathsfit{#1}}}

\def\tW{{\tens{W}}}

\DeclareMathOperator*{\argmax}{arg\,max}
\DeclareMathOperator*{\argmin}{arg\,min}

%% file: sections/abstract.tex
\begin{abstract}
	The emerging paradigm of federated learning strives to enable collaborative training of machine learning models on the network edge without centrally aggregating raw data and hence, improving data privacy.
	This sharply deviates from traditional machine learning and necessitates design of algorithms robust to various sources of heterogeneity.
	Specifically, statistical heterogeneity of data across user devices can severely degrade performance of standard federated averaging for traditional machine learning applications like personalization with deep learning.
	This paper proposes \FedPerAlg{}, a base + personalization layer approach for federated training of deep feed forward neural networks, which can combat the ill-effects of statistical heterogeneity.
	We demonstrate effectiveness of \FedPerAlg{} for non-identical data partitions of \cifar{} datasets and on a personalized image aesthetics dataset from Flickr.
\end{abstract}


%% file: sections/introduction.tex
\section{Introduction}
	\label{sec:intro}
	Modern day mobiles, wearables, home assistants and other internet enabled devices possess sophisticated hardware and software capabilities to support a variety of applications and generate large volumes of contextual and user data.
	Owing to privacy concerns and increasing compute and storage capabilities at the network edge, it has become increasingly attractive to keep data locally on user/client/edge devices and execute machine learning (ML) model training computations on device with locally available data and occassional communication with an aggregating parameter server.
	This approach to training ML models is termed as \emph{federated learning} \citep{McMahanMooreRamageEtAl2017CommunicationEfficientLearning} and is in sharp contrast to traditional ML training and prediction pipelines that rely on central aggregation of raw data.

	The recent overview article by \citet{LiSahuTalwalkarEtAl2019FederatedLearningChallenges} articulates the many unique challenges faced by a federated learning system.
	One such challenge is that the effective data distribution at different clients can vary greatly across the (potentially millions) of participating devices.
	Such \emph{statistical heterogeneity} can be detrimental to the performance of ML training algorithms for several applications like personalization, recommendation, fraud detection, \etc~since traditional ML training algorithms are designed for central or distributed computation environments where data partitioning can be tightly controlled.
	Overcoming the ill-effects of statistical heterogeneity in federated learning is an active area of research with several recent works~\citep{ChenDongLiEtAl2018FederatedMetaLearning,SmithChiangSanjabiEtAl2017Federatedmultitask,ZhaoLiLaiEtAl2018FederatedLearningNon,SahuLiSanjabiEtAl2018ConvergenceFederatedOptimization}.

	Herein, we study the effects of the personalization setup as a source of statistical heterogeneity in federated learning with deep feedforward neural networks.
	Personalization is a key application of ML and is possible because users differ in their preferences as captured from raw user data.
	In a federated setup where edge devices belong to users, this necessarily means that data for personalization is statistically heterogeneous.
	For personalization via recommendation, \citet{Ammaduddin2019FederatedCF} present the first attempt to extend collaborative filtering to a federated setup and \citet{ChenDongLiEtAl2018FederatedMetaLearning} design a meta-learning approach for federated training.
	While these approaches were demonstrated to perform very well on the MovieLens 100k dataset \citep{HarperKonstan2015MovieLensDatasetsHistory}, it isn't obvious how to extend them to deep neural network models or to problems that cannot be addressed by collaborative filtering.
	We believe that the right approach to personalized federated learning is a highly non-trivial question on which the research community has barely scratched the surface.
	In particular, the challenge comes from multiple facets, including but not limited to:
	\begin{itemize}
		\item	Many personalization tasks like personalized image aesthetics \citep{RenShenLinEtAl2017PersonalizedImageAesthetics} and personalized highlight detection \citep{GarciadelMolinoGygli2018PHDGIFsPersonalized} have no explicit user features in the data and need to extract such features during training.
		Hence, same input data can receive different labels from different users implying that the personalized models must differ across users to be able to predict different labels for similar test data.
		This is already outside the scope of standard federated learning~\citep{McMahanMooreRamageEtAl2017CommunicationEfficientLearning} which learns one global model and effectively replicates it locally on every client.
		\item	The number of training samples per user is not large enough to train individual ML models in isolation.
		Hence, some communication and collaboration is needed to leverage \emph{wisdom of the crowd}.
		However, collaborative filtering algorithms may not be applicable since they require considerable overlap between items rated by different users.
		According to \citet{RenShenLinEtAl2017PersonalizedImageAesthetics,GarciadelMolinoGygli2018PHDGIFsPersonalized}, the sufficient overlap condition fails in typical datasets for personalized image aesthetics and personalized highlight detection among other tasks.
	\end{itemize}

	\begin{figure}[t]
		\centering
		\includegraphics[width=\figwidth]{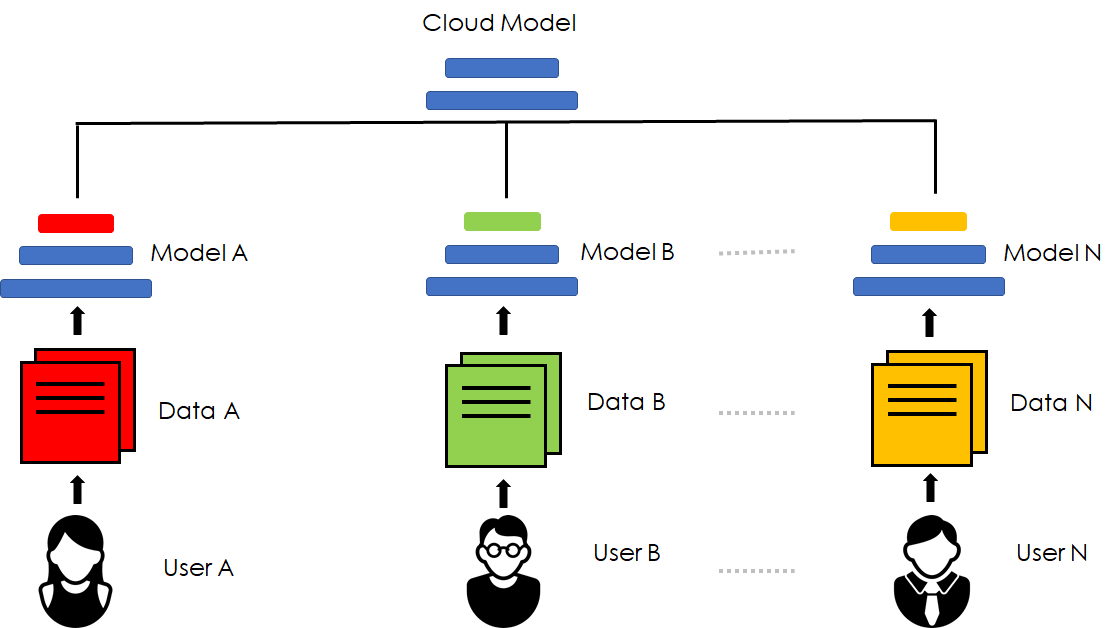}
		\caption{Pictorial view of proposed federated personalization approach. All user devices share a set of base layers with same weights (colored blue) and have distinct personalization layers that can potentially adapt to individual data. The base layers are shared with the parameter server while the personalization layers are kept private by each device.}
		\label{fig:model}
	\end{figure}

	\subsection{Contributions}
	\begin{enumerate}
		\item	We propose to capture personalization aspects in federated learning by viewing deep learning models as \emph{base} + \emph{personalization} layers as illustrated in \figurename~\ref{fig:model}.
		Our training algorithm comprises of the base layers being trained by federated averaging (or some variant thereof) and personalization layers being trained only from local data with stochastic gradient descent (or some variant thereof).
		We demonstrate that the personalization layers that are free from the federated averaging (\FedAvgAlg) procedure can help combat the ill-effects of statistical heterogeneity.
		\item	We demonstrate that standard federated learning setup is highly unsuitable for personalization tasks by comparing its performance against our federated personalization approach \wrt~two datasets:
		\begin{enumerate*}
			\item	non-identically partitioned \cifar[10]/\cifar[100] dataset, and
			\item	\flickr{} dataset from \citet{RenShenLinEtAl2017PersonalizedImageAesthetics}.
		\end{enumerate*}
		Interestingly, standard federated learning's failure on the \flickr{} dataset is qualitatively and quantitatively very different from its poor performance on the non-identically partitioned \cifar[100] dataset.
	\end{enumerate}
	
	\bibsubfile{apalike}{bibfile}

%% file: sections/related-work.tex
\section{Related Work}
	\label{sec:related}
	We have built upon several ideas in literature for the research communicated in this paper.
	We call out the overlaps and differences below.
	
	\textbf{Federated Learning:}
		This paradigm for training ML models on user/client/edge devices was first proposed through the papers \citet{KonecnyMcMahanRamage2015FederatedOptimizationDistributed,McMahanMooreRamageEtAl2017CommunicationEfficientLearning,KonecnyMcMahanYuEtAl2018FederatedLearningStrategies}.
		The initial focus was on reducing the communication overhead while maintaining the statistical performance of the learned global model.
		Since these early works, researchers have worked to shed light on the other challenges of making a federated system work (which include \emph{system heterogeneity} and \emph{privacy leakage} besides \emph{statistical heterogeneity} and \emph{communication overhead}) and potential solutions in various application scenarios.
		These are highlighted with bibliographic references in the recent overview article \citet{LiSahuTalwalkarEtAl2019FederatedLearningChallenges}.
		While we acknowledge that progress on all four challenges are needed for practically realizing federated learning systems, \emph{our novelty in this work is restricted to the statistical heterogeneity aspect while maintaining composability with the current approaches to tackle the other three challenges}.
		Our `base + personalization layers' notion for federated deep learning effectively learns different overlapping local models to better capture statistical heterogeneity.
		This approach expands on the original scope of federated learning which was to train one global model and essentially replicate it on all user devices.
		It is known that the original \FedAvgAlg{} algorithm is possibly non-convergent when training on pathological and/or highly non-identical data partitions \citep{LiSahuTalwalkarEtAl2019FederatedLearningChallenges,McMahanMooreRamageEtAl2017CommunicationEfficientLearning}.

	\textbf{Multi-task and Transfer Learning:}
		Modeling and algorithmic techniques in multi-task and transfer learning domains try to learn with lesser training data by leveraging latent relationships across multiple tasks defined on the same dataset.
		\citet{SmithChiangSanjabiEtAl2017Federatedmultitask} were the first to recognize that multi-task learning models could be effectively used to study various heterogeneities of a federated learning setup and incorporate aspects of personalization of local models.
		The authors restrict consideration to convex/bi-convex formulations of multi-task learning which cannot be easily extended to neural network models.
		Since deep learning models naturally yield to multi-task formulations due to their layered structure \citep{Ruder2017OverviewMultiTask}, it should be possible to study federated deep multi-task learning setups that generalize assumptions in \citet{SmithChiangSanjabiEtAl2017Federatedmultitask}.
		This is indeed the primary inspiration for the `base + personalization layers' formulation proposed in this paper.
		Intuitively, the base layers can act as the shared layers in a multi-task learning model and capture the wisdom of the crowd, while the personalization layers can act as the task specific layers of the same multi-task learning model to capture user specific aspects.

		In a separate line of work \citep{ChenWangYuEtAl2019FedHealthFederatedTransfer,VepakommaGuptaSwedishEtAl2018Splitlearninghealth} motivated by scenarios in the healthcare domain, a federated transfer learning approach has been proposed for solving the personalization problem.
		A hidden assumption in this approach is that it is possible to separate the global training steps from the subsequent local personalization steps.
		We don't see any intuitive reason for such an assumption to hold for use cases outside of the healthcare domain and it definitely does not hold for the personalized image aesthetics and the personalized highlight detection tasks in \citet{RenShenLinEtAl2017PersonalizedImageAesthetics,GarciadelMolinoGygli2018PHDGIFsPersonalized}.
		In the healthcare domain, separate organizations can be thought of as `devices' that are disallowed from sharing raw data but could jointly train ML models via federated learning techniques \citep{YangLiuChenEtAl2019FederatedMachineLearning}.
		\citet{VepakommaGuptaSwedishEtAl2018Splitlearninghealth} motivates various ways to split a federated deep learning model across organizational boundaries to satisfy different constraints on data sharing for vertically partitioned data.
		Although our `base + personalization layers' construct is also based on a \emph{splitting} principle, our specific split is different from those considered in the healthcare domain, and is motivated from and evaluated on completely different datasets and much smaller data volume distributions.

	\textbf{Distributed Personalization and Recommendation:}
		Centralized algorithms for personalization and recommendation and their distributed counterparts have been extensively studied for various applications.
		The key assumption for many distributed ML algorithms is that data partitioning can be tightly controlled and communication delays are \iid{}
		Remarkably, even asynchronous distributed algorithms like in \cite{MiaoChuTangEtAl2015DistributedPersonalization} can be shown to enjoy strong theoretical and empirical properties (like sublinear convergence for convex losses) under these assumptions.
		However, careful redesign and analysis of personalization and recommendation algorithms is needed for the federated setup owing to statistical and communication heterogeneity.
		Such efforts have appeared only recently in the form of \citet{Ammaduddin2019FederatedCF} developing the first federated collaborative filtering algorithm and \citet{ChenDongLiEtAl2018FederatedMetaLearning} formulating a federated meta-learning approach for recommendation.
		At present, these efforts disregard deep learning models, which is a significant shortcoming.

	\bibsubfile{apalike}{bibfile}

%% file: sections/algorithm.tex
\section{Modeling and Algorithmic Setup}
	\label{sec:algorithm}
	\subsection{Model}
		Consider the setup shown in \figurename~\ref{fig:model} where all user devices share the same base layers and have unique personalization layers constituting the deep feed forward neural network models.
		We denote the number of base and personalized layers on each client by positive integers $K_B$ and $K_P$ respectively and let there be $N$ user devices.
		Let us designate the base layer weight matrices by $\mat{W}_{B,1}, \mat{W}_{B,2}, \dotsc, \mat{W}_{B,K_B}$ and their corresponding vector-valued activation functions as $\vec{a}_{B,1}, \vec{a}_{B,2}, \dotsc, \vec{a}_{B,K_B}$.
		While the base layer weight matrices may be of different dimensions, with a slight abuse of notation we will designate the tuple $\bb{\mat{W}_{B,K_B}, \dotsc, \mat{W}_{B,2}, \mat{W}_{B,1}}$ as $\tW_B$, using the tensor notation.
		Analogously, for $j \in \cc{1, 2, \dotsc, N}$, we let the personalized layer weight matrices for the $j^{\thp}$ user device be designated as $\mat{W}_{P_j,1}, \mat{W}_{P_j,2}, \dotsc, \mat{W}_{P_j,K_P}$, the corresponding vector-valued activation functions be denoted by $\vec{a}_{P_j,1}, \vec{a}_{P_j,2}, \dotsc, \vec{a}_{P_j,K_P}$, and the tuple $\bb{\mat{W}_{P_j,K_P}, \dotsc, \mat{W}_{P_j,2}, \mat{W}_{P_j,1}}$ be notationally captured by $\tW_{P_j}$.
		We will not use any non-trivial tensor calculus in the paper so the abuse of notation should not cause any confusion.
		Further, we will not explicitly track the dimensions of any weight matrices as the dimensions are irrelevant for the exposition in this paper.
		The dimensions of the domain and range of the vector valued activation functions are implicitly unique so as to satisfy the forward pass/inference operation
		\makeatletter
			\if@twocolumn
				\begin{multline}
					\hat{y} = \vec{a}_{P_j,K_P}\bd[\big]{\mat{W}_{P_j,K_P} \cdots \vec{a}_{P_j,1}\bd[\big]{\mat{W}_{P_j,1} \vec{a}_{B,K_B}\bd[\big]{}}}	\\
					\db[\big]{\db[\big]{\db[\big]{\mat{W}_{B,K_B} \cdots \vec{a}_{B,1}\bb{\mat{W}_{B,1} \vec{x}} \cdots}} \cdots}
					\label{eqn:forward-pass}
				\end{multline}
			\else
				\begin{equation}
					\hat{y} = \vec{a}_{P_j,K_P}\bb{\mat{W}_{P_j,K_P} \cdots \vec{a}_{P_j,1}\bb{\mat{W}_{P_j,1} \vec{a}_{B,K_B}\bb{\mat{W}_{B,K_B} \cdots \vec{a}_{B,1}\bb{\mat{W}_{B,1} \vec{x}} \cdots}} \cdots}
					\label{eqn:forward-pass}
				\end{equation}
			\fi
		\makeatother
		for any input data point $\vec{x}$ at the $j^{\thp}$ device.
		For brevity, we'll denote this forward pass operation at the $j^{\thp}$ device by $\hat{y} = f\bb{\vec{x}; \tW_B, \tW_{P_j}}$.
		Thus, we implicitly assume that the weight tensor $\tW_{P_j}$ captures all aspects of personalization at the $j^{\thp}$ device.
	
		With a slight abuse of notation, we denote the joint distribution that generates (data, label) pairs at the $j^{\thp}$ device by $P_j$.
		Letting $l\bb{\cdot,\cdot}$ denote the per sample loss function common to all devices, the learning goal in our setting is to minimize the average personalized population risk function
		\makeatletter
			\if@twocolumn
				\begin{multline}
					L^{PR}\bb[\big]{\tW_B, \tW_{P_1}, \dotsc, \tW_{P_N}}	\\
					= \frac{1}{N}\sum_{j=1}^{N} \expect<\bb{\rvx,\ry} \sim P_j> {l\bb{\ry, f\bb{\rvx; \tW_B, \tW_{P_j}}}}
				\end{multline}
			\else
				\begin{equation}
					L^{PR}\bb[\big]{\tW_B, \tW_{P_1}, \dotsc, \tW_{P_N}} = \frac{1}{N}\sum_{j=1}^{N} \expect<\bb{\rvx,\ry} \sim P_j> {l\bb{\ry, f\bb{\rvx; \tW_B, \tW_{P_j}}}}
				\end{equation}
			\fi
		\makeatother
		with respect to the weight tensors $\tW_B, \tW_{P_1}, \dotsc, \tW_{P_N}$.
		Since the true data generating distributions $P_j$ are unknown during training, we'll use the empirical risk at the $j^{\thp}$ device
		\begin{equation}
			L^{ER}_{j}\bb[\big]{\tW_B, \tW_P} \triangleq \frac{1}{n_j}\sum_{i=1}^{n_j} l\bb{y_{j,i}, f\bb{\vec{x}_{j,i}; \tW_B, \tW_P}}
			\label{eqn:client-empirical-risk}
		\end{equation}
		as a proxy for the corresponding population risk $\expect<\bb{\rvx,\ry} \sim P_j>{l\bb{\ry, f\bb{\rvx;\tW_B,\tW_P}}}$, where $n_j$ is the number of training samples available at the $j^{\thp}$ device and $\bb{\vec{x}_{j,i}, y_{j,i}}$ is the $i^{\thp}$ such sample for $i \in \cc{1, \dotsc, n_j}$.

	\subsection{Algorithm}
		We will use stochastic gradient descent (\sgd) as a subroutine in our proposed algorithm.
		Standard formulations of \sgd{} \citep{BottouCurtisNocedal2018OptimizationMethodsLarge} for minimizing an empirical risk function like \eqref{eqn:client-empirical-risk} requires us to specify the following:
		\begin{enumerate*}[(a)]
			\item	the decision variables that would be updated by \sgd{} and their initial values,
			\item	the batch size for partitioning the dataset $\set{\bb{\vec{x}_{j,i}, y_{j,i}}}{1 \leq i \leq n_j}$ and the number of epochs over that dataset, and
			\item	the learning rate.
		\end{enumerate*}
		We will denote by $\sgd\bb[\big]{L^{ER}_{j}, \widehat{\tW}_B, \widehat{\tW}_P, \eta, e, b}$, the updated values of decision variable $\bb[\big]{\tW_B, \tW_P}$, when the \sgd{} is started from $\bb[\big]{\tW_B, \tW_P} = \bb[\big]{\widehat{\tW}_B, \widehat{\tW}_P}$ \wrt~the loss function $L^{ER}_{j}\bb[\big]{\tW_B, \tW_P}$ and executed with learning rate $\eta$ and batch size $b$ for $e$ epochs over the dataset.
		We have assumed that the random permutation of the dataset for an epoch is defined independent of the proposed algorithm.
		
		Our proposed personalized federated training algorithm \FedPerAlg{} is described via \algorithmsname~\ref{alg:FedPerClient} and \ref{alg:FedPerServer}.
		The steps for the server component of \FedPerAlg{} are detailed under \algorithmname~\ref{alg:FedPerServer} and the steps for the $j^{\thp}$ client are described under \algorithmname~\ref{alg:FedPerClient}.
		Intuitively, the server employs a \FedAvgAlg{} based approach to train the base layers globally whereas each client updates its base and personalized layers locally (between successive global aggregations) using a \sgd{} style algorithm.
		To keep the exposition simple and relevant to the personalization driven statistical heterogeneity, we make the following assumptions that can be readily relaxed in implementation:
		\begin{enumerate*}[(a)]
			\item	the dataset at any client doesn't change across global aggregations,
			\item	the batch size $b$ and the \#epochs $e$ are invariant across clients and across global aggregations,
			\item	each client uses \sgd{} to update $\bb{\tW_B, \tW_{P_j}}$ between global aggregations, and
			\item	all $N$ user devices are active throughout the training process.
		\end{enumerate*}
		
		We denote the values assumed during the $k^{\thp}$ iteration by $\bb{k}$ in the superscript.
		Between the $k^{\thp}$ and $\bb{k+1}^{\thp}$ global aggregations, the $j^{\thp}$ device execute \sgd{} with learning rate $\eta_j^{\bb{k}}$ and let the updated value $\sgd\bb[\big]{L^{ER}_{j}, \widehat{\tW}_B, \widehat{\tW}_P, \eta_j^{\bb{k}}, e, b}$ be denoted by the shorthand $\sgd_j\bb[\big]{\widehat{\tW}_B, \widehat{\tW}_P, \eta_j^{\bb{k}}}$.
		During the $k^{\thp}$ global aggregation, the server computes a weighted combination of $\tW_{B,j}^{\bb{k}}$ across all clients $j \in \cc{1,\dotsc,N}$ using weights $\gamma_j = n_j/\sum_{j=1}^{N} n_j$ to arrive at $\tW_B^{\bb{k}}$.

		\begin{algorithm}[t]
			\caption{\FedPerAlg-\textsc{Client}($j$)}
			\begin{algorithmic}[1]
				\Require	$f\bb{\cdot; \cdot, \cdot}, e, b, \set{\bb{\vec{x}_{j,i}, y_{j,i}}}{i \in \cc{1, \dotsc, n_j}}$
				\Require	$\eta_j^{\bb{k}}$ for $k \in \setZ_{+}$
				\State	Initialize $\tW_{P_j}^{\bb{0}}$ at random
				\State Send $n_j$ to server
				\For{$k = 1,2,\dots$}
					\State	Receive $\tW_B^{\bb{k-1}}$ from server
					\State	$\bb[\big]{\tW_{B,j}^{\bb{k}}, \tW_{P_j}^{\bb{k}}} \gets \sgd_j\bb[\big]{\tW_B^{\bb{k-1}}, \tW_{P_j}^{\bb{k-1}}, \eta_j^{\bb{k}}}$
					\State	Send $\tW_{B,j}^{\bb{k}}$ to server
				\EndFor
			\end{algorithmic}
			\label{alg:FedPerClient}
		\end{algorithm}
	
		\begin{algorithm}[t]
			\caption{\FedPerAlg-\textsc{Server}}
			\begin{algorithmic}[1]
				\State	Initialize $\tW_B^{\bb{0}}$ at random
				\State	Receive $n_j$ from each client $j \in \cc{1,\dotsc,N}$ and compute $\gamma_j = n_j/\sum_{j=1}^{N} n_j$
				\State	Send $\tW_B^{\bb{0}}$ to each client
				\For{$k = 1,2,\dots$}
					\State	Receive $\tW_{B,j}^{\bb{k}}$ from each client $j \in \cc{1,\dotsc,N}$
					\State	Aggregate $\tW_B^{\bb{k}} \gets \sum_{j=1}^{N} \gamma_j\tW_{B,j}^{\bb{k}}$
					\State	Send $\tW_B^{\bb{k}}$ to each client
				\EndFor
			\end{algorithmic}
			\label{alg:FedPerServer}
		\end{algorithm}

\bibsubfile{apalike}{bibfile}

%% file: sections/experiments.tex
\section{Experiments}
\label{sec:experiments}
	
	    
	 

	\begin{figure*}
		\centering
		\begin{subfigure}[b]{0.45\textwidth}
			\centering
			\includegraphics[width=\textwidth]{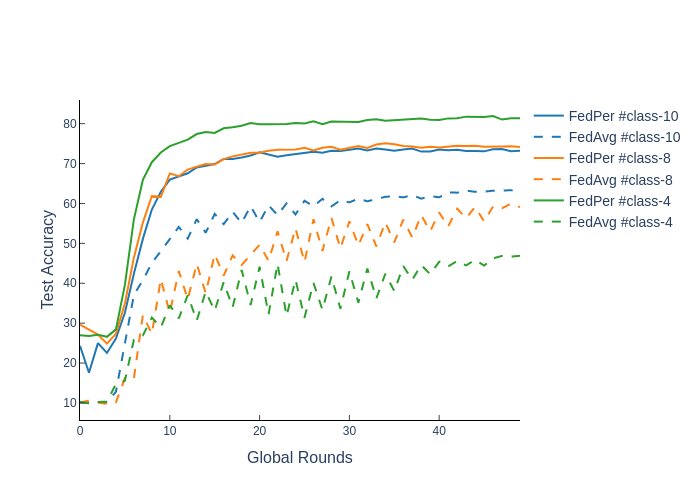}
			\caption{\mobnet[v1]}
			\label{fig:mobilenetoverlapping10}
		\end{subfigure}
		\hfill
		\begin{subfigure}[b]{0.45\textwidth}
			\centering
			\includegraphics[width=\textwidth]{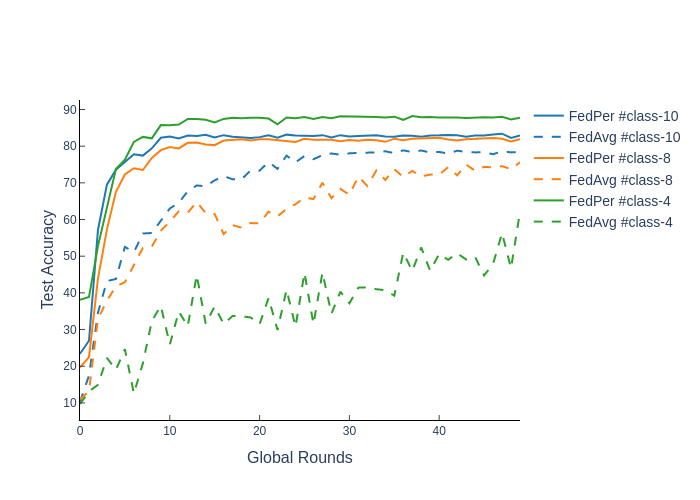}
			\caption{\resnet[34]}
			\label{fig:resnetoverlapping10}
		\end{subfigure}
		\caption{Performance of \FedAvgAlg{} vs \FedPerAlg{} on non-identical \cifar[10] partition ($k \in \cc{4,8,10}$)}
		\label{fig:overlapping10}
	\end{figure*}
%

	\begin{figure}[t]
		\centering
		\includegraphics[width=\figwidth]{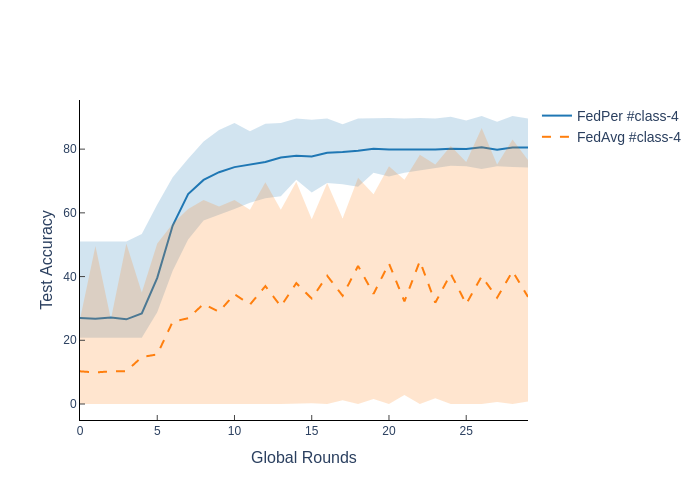}
		\caption{Variation in \mobnet[v1] performance across clients for \FedAvgAlg{} vs \FedPerAlg{} on non-identical \cifar[10] partition ($k=4$)}
		\label{fig:mobilenetoverlappingvariation4}
	\end{figure}

	\begin{figure*}
		\centering
		\begin{subfigure}[b]{0.45\textwidth}
			\centering
			\includegraphics[width=\textwidth]{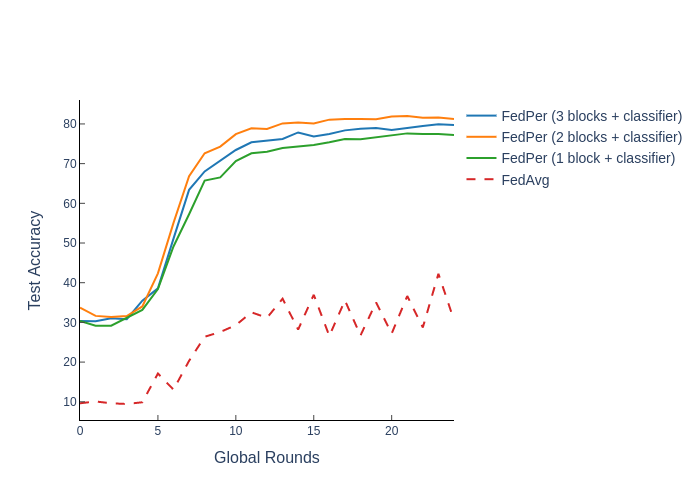}
			\caption{\mobnet[v1]}
			\label{fig:mobilenet_personal_cifar10}
		\end{subfigure}
		\hfill
		\begin{subfigure}[b]{0.45\textwidth}
			\centering
			\includegraphics[width=\textwidth]{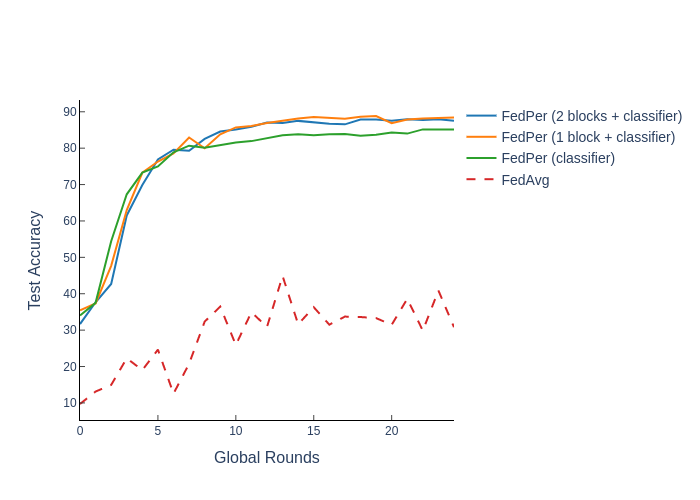}
			\caption{\resnet[34]}
			\label{fig:resnet_personal_cifar10}
		\end{subfigure}
		\caption{Performance of \FedPerAlg{} on non-identical \cifar[100] partition \wrt~\#personalization layers}
		\label{fig:personal_cifar10}
	\end{figure*}

%

We evaluate the performance of \FedPerAlg{} across multiple datasets and deep learning model families and contrast it with the behavior of \FedAvgAlg{}.
We note that \FedPerAlg{} reduces to \FedAvgAlg{} when $K_P = 0$, \ie~personalization layers are absent.
\sectionsname~\ref{sec:exp:stat-het} and \ref{sec:exp:per-layers} respectively present the effect of statistical heterogeneity and changing $K_P$ for experiments with \cifar{} datasets.
Corresponding results on the \flickr{} dataset are presented in \sectionname~\ref{sec:exp:img-aes}.
In the spirit of reproducibility, we have made all datasets, implementations for experiments, and results available at \url{https://bit.ly/35dKebE}.

\subsection{Datasets, Model Architectures, and Implementation Details}
	\textbf{Datasets:}
	\begin{enumerate}
		\item	\cifar[10] and \cifar[100] are image classification datasets with 10 and 100 labeled classes respectively \citep{krizhevsky2009learning}.
		Both have 50,000 training samples and 10,000 testing samples.
		Each image has a unique label.
		\item	\flickr{} dataset was compiled by \citet{RenShenLinEtAl2017PersonalizedImageAesthetics} to study personalized image aesthetics.
		It has 40,000 photographs from \url{www.flickr.com} with aesthetic ratings (between 1 and 5) collected via Amazon Mechanical Turk.
		Each image is rated by 5 different users and there are a total of 210 users.
		Each user has rated between 60 and 290 images with average being 160 ratings.
		We randomly split the data of each user into an 80\% training set and a 20\% testing set.
	\end{enumerate}

	\textbf{Model Architectures:}
	We work with the \resnet[34] and \mobnet[v1] families of convolution neural networks.
	While \resnet{} model family achieves state-of-the-art performance for many image classification tasks, \mobnet{} is an efficient model family for mobile vision applications with low compute resources.
	Both of these families are constructed by composing their respective basic block multiple times.
	\resnet[34] and \mobnet[v1] have 16 and 11 basic blocks respectively.
	\resnet's basic block is a residual block consisting of 2 convlayers and a residual connection, while \mobnet's basic block consists of a depthwise and a pointwise convolutional layer.
	We will delineate base and personalization layers in units of basic blocks.

	\textbf{Implementation Details:}
	We run each experiment for 100 global aggregations, with $e = 4$ epochs for \sgd{} between successive global aggregations.
	We use constant learning rate $\eta_j^{\bb{k}} = 0.01$ across global aggregations and across clients.
	\begin{itemize}
		\item	Experiments with \flickr{} dataset use a random subset of $N = 30$ users as clients, with \sgd{} batch size $b = 4$.
		This dataset is naturally unbalanced in terms of $n_j$ and statistically heterogeneous in terms of $P_j$.
		\item	Experiments with \cifar[10] and \cifar[100] use $N = 10$ clients, with \sgd{} batch size $b = 128$.
		Data volumes are balanced across clients in terms of $n_j$, but are non-identically partitioned in terms of $P_j$ by restricting the samples of any particular client to belong to atmost $k$ classes.
		The number of samples from a particular class is distributed equally across clients allowed to sample from it, and the level of statistical heterogeneity is varied by changing the value of $k$.
	\end{itemize}

\subsection{Effect of Statistical Heterogeneity}
	\label{sec:exp:stat-het}
	\figurename~\ref{fig:overlapping10} presents the test set accuracies for \FedAvgAlg{} vs \FedPerAlg{} (averaged across clients) on \resnet[34] and \mobnet[v1] as a function of global aggregation rounds for different levels of statistical heterogeneity ($k \in \cc{4,8,10}$).
	$k = 4$ corresponds to a highly non-identical data partition, whereas $k = 10$ corresponds to an identical data partition across the clients.
	\figurename~\ref{fig:mobilenetoverlappingvariation4} shows the variation of the test set accuracies across clients (via the shaded regions) for \FedAvgAlg{} vs \FedPerAlg{} for $k = 4$ with \mobnet[v1].
	For results in both of these figures, the classifier layer (final fully connected layer) and the immediately preceding basic block were used as personalization layers for \FedPerAlg.
	
	We observe from \figurename~\ref{fig:overlapping10} that \FedPerAlg{} is significantly better both in terms of convergence speed \wrt~global rounds and the client averaged test accuracy achieved at steady state.
	Interestingly, as the data partition becomes more identical (parameterized by $k$ moving from 4 to 10), \FedPerAlg's performance comes closer to that of \FedAvgAlg.
	The qualitative nature of these results remains unchanged even if \cifar[10] is replaced by \cifar[100] and we select $k \in \cc{40,80,100}$ (see supplementary material for plots).

	\figurename~\ref{fig:mobilenetoverlappingvariation4} shows that \FedPerAlg{} also results in lower variation in test accuracies across clients.
	This is important for fairness in learning.
	The large cross-client variation in test accuracies for \FedAvgAlg{} is somewhat surprising.
	However, the magnitude of this variation does decrease when non-identicality of the data partition is reduced by increasing $k$ from 4 to 10 (see supplementary material for plots).

\subsection{Effect of Personalization Layers}
	\label{sec:exp:per-layers}
	We consider personalization layers to include the classifier layer (final fully connected layer) and last few basic blocks.
	Since we delineate personalization layers in units of basic blocks, with a slight abuse of notation, we can use $K_P$ to denote the number of basic blocks included in the personalization layers.
	$K_P = 4$ refers to the classifier layer and the last 3 basic blocks included in the set of personalization layers, whereas $K_P = 1$ refers to only the classifier layer included in the set of personalization layers.
	
	\figurename~\ref{fig:personal_cifar10} plots test accuracies (averaged across clients) \wrt~\cifar[10] for \FedPerAlg{} on \resnet[34] and \mobnet[v1] as a function of global aggregation rounds for different number of personalization layers ($K_P \in \cc{1,2,3,4}$).
	Interestingly, there doesn't seem to be a clear correlation between $K_P$ and the client averaged test accuracy at steady state, except that having atleast one personalization layer helps.
	For both \mobnet[v1] and \resnet[34], $K_P = 2$ seems to achieve the best performance on \cifar[10], but $K_P = 1$ seems to be the best for \cifar[100].

\subsection{Do Base Layers Learn Anything?}
\begin{figure}[t]
  \centering
  \includegraphics[width=\figwidth]{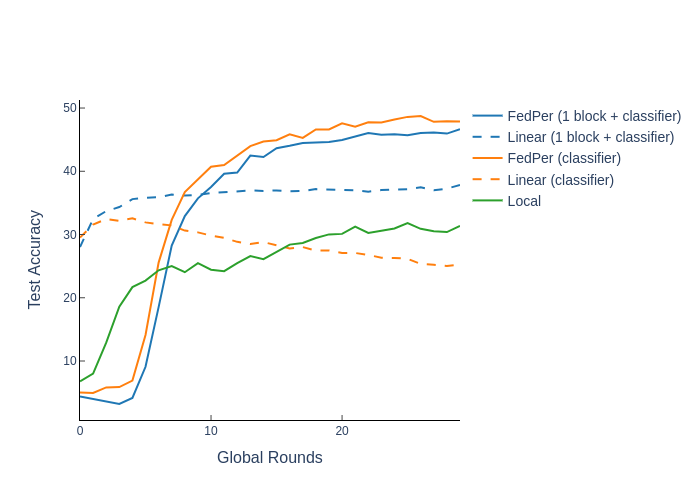}
  \caption{Effect on the performance of \mobnet[v1] on \cifar[100] before and after replacing the base layers with linear layers}
  \label{fig:MobileNet_identity_cifar100}
\end{figure}

\begin{figure}[t]
  \centering
  \includegraphics[width=\figwidth]{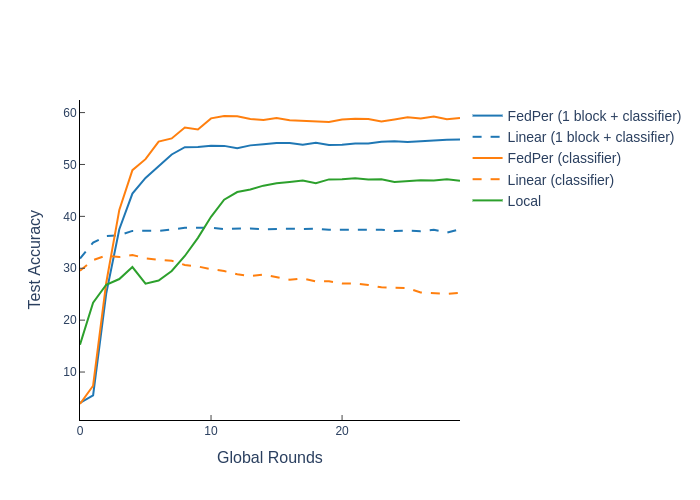}
  \caption{Effect on the performance of \resnet[34] on \cifar[100] before and after replacing the base layers with linear layers}
  \label{fig:Resnet_identity_cifar100}
\end{figure}

	In theory, it is possible that the personalization layers may be sufficient for the learning tasks at hand rendering the base layers redundant.
	To test whether this is the case, we replace the base layers at each client with a linear fully connected layer and train with \FedPerAlg.
	If the base layers are redundant, then there should not be a significant drop in test accuracy (as compared to \resnet[34] or \mobnet[v1]) when this model is trained.
	\figuresname~\ref{fig:MobileNet_identity_cifar100} and \ref{fig:Resnet_identity_cifar100} confirm that this is not the case, atleast for \cifar[100].
	In fact, the performance of purely local training confirms that the number of training samples per client is not large enough to learn individual client models to high degrees of accuracy.
  
\subsection{Performance on \flickr}
	\label{sec:exp:img-aes}

\begin{figure}[t]
  \centering
  \includegraphics[width=\figwidth]{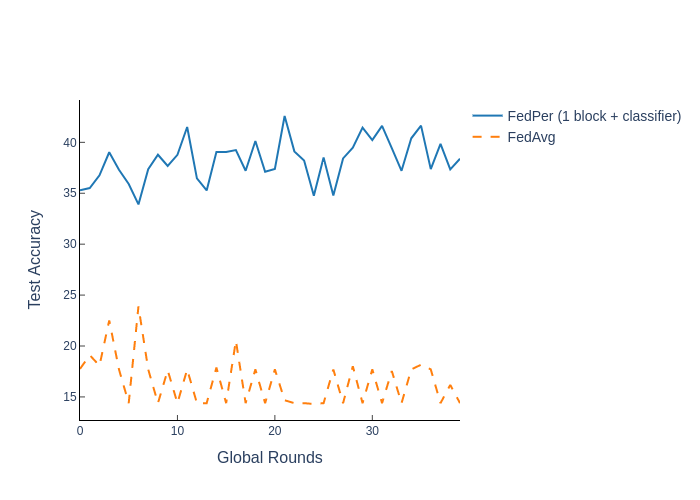}
  \caption{Performance of \mobnet[v1] on \flickr{} when trained using \FedAvgAlg{} and \FedPerAlg{}}
  \label{fig:mobilenet_aes}
\end{figure}

\begin{figure}[t]
  \centering
  \includegraphics[width=\figwidth]{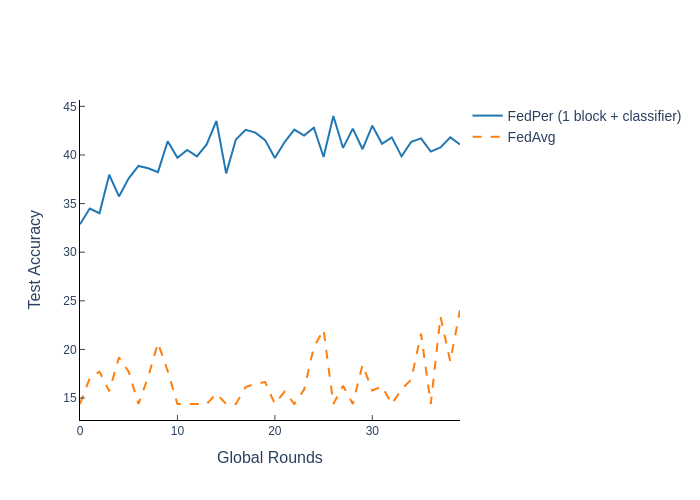}
  \caption{Performance of \resnet[34] on \flickr{} when trained using \FedAvgAlg{} and \FedPerAlg{}}
  \label{fig:resnet_aes}
\end{figure}

\figurename~\ref{fig:mobilenet_aes} and \ref{fig:resnet_aes} shows the test performance of the \mobnet[v1] and \resnet[34] models on the \flickr{} dataset.
It can be seen from the results that \FedAvgAlg{} is unfit for modeling personalization tasks.
The Performance of the \mobnet{} model trained using \FedAvgAlg{} is quantitatively very different when compared to its performance on \cifar{} datasets and is similar to random guessing.
This is because, the \FedAvgAlg{} approach has no way of capturing the personal preferences of the user as it has the same copy of the model on all clients.

\begin{figure}[t]
  \centering
  \includegraphics[width=\figwidth]{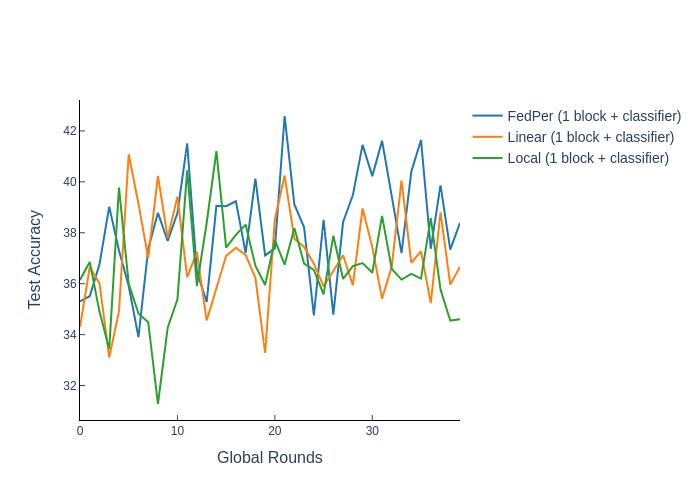}
  \caption{Effect on the performance of \mobnet[v1] on \flickr{} before and after replacing the base layers with linear layers}
  \label{fig:mobilenet_aes_identity}
\end{figure}

Models trained using our \FedPerAlg{} approach captures the user preferences through the personalization layers.
While the models used here are not state-of-the-art for this task, it provides sufficient evidence to show that our \FedPerAlg{} approach has the ability to model personalized tasks.
Results of the experiment where we replace the base layers of the model with linear layer for \mobnet[v1] is given in \figurename~\ref{fig:mobilenet_aes_identity}.
Similar to the \cifar[100] dataset, we see a drop in the performance of morphed models thereby again confirming our hypothesis that base layers capture complex image features through federated training.

\begin{figure}[t]
  \centering
  \includegraphics[width=\figwidth]{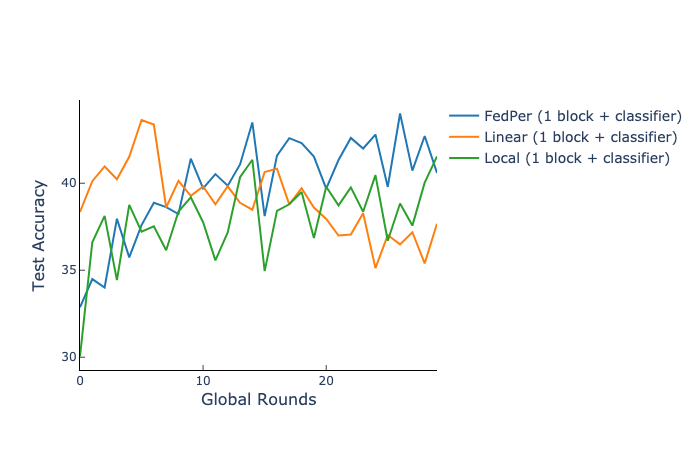}
  \caption{Effect on the performance of \resnet[34] on \flickr{} before and after replacing the base layers with linear layers}
  \label{fig:resnet_aes_identity}
\end{figure}

\bibsubfile{apalike}{bibfile}

%% file: sections/conclusions.tex
\section{Conclusion}
	\label{sec:conclude}
	In this work, we propose \FedPerAlg{} a novel approach for capturing the personalization aspect of users in the federated learning setup.
	\FedPerAlg{} achieves this by viewing the deep learning models as base + personalization layers.
	While the base layers are trained in a collaborative manner using the existing federated learning approach, the personalization layers are trained locally thereby enabling our approach to capture personalization aspects of users.
	Results indicate that having base + personalization layers help in combating the ill-effects of statistical heterogeneity.
	Empirical results on \flickr{} and \cifar{} datasets demonstrates the ineffectiveness of \FedAvgAlg{} and the effectiveness of \FedPerAlg{} in modelling the personalization tasks.


%% file: sections/appendix.tex
\appendix
\section{Appendix}
	\label{sec:appendix}
	\subsection{Effect of Statistical Heterogeneity}
\begin{figure}[t]
		\centering
		\includegraphics[width=\figwidth]{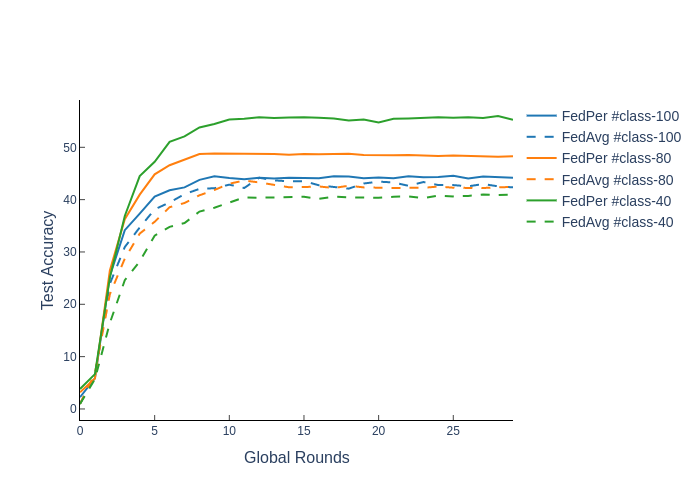}
		\caption{Effect of Non-IID nature on the performance of \resnet[34] on \cifar[100] when trained using \FedAvgAlg{} and \FedPerAlg{}}
		\label{fig:resnetoverlapping100}
	\end{figure}

\begin{figure}[t]
		\centering
		\includegraphics[width=\figwidth]{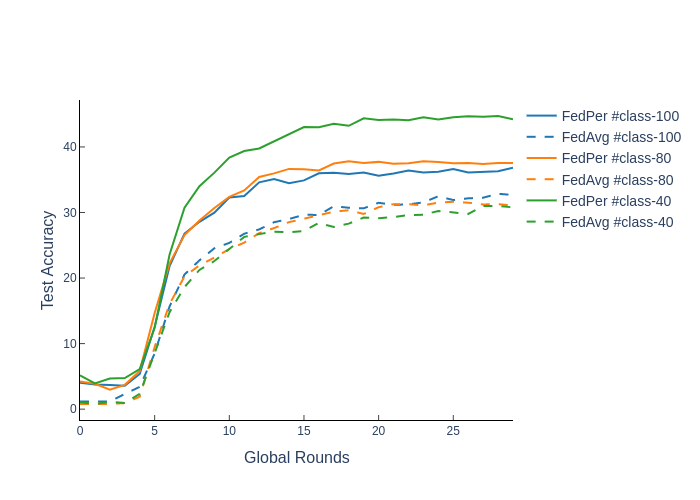}
		\caption{Effect of Non-IID nature on the performance of \mobnet[v1] on \cifar[100] when trained using \FedAvgAlg{} and \FedPerAlg{}}
		\label{fig:mobilenetoverlapping100}
	\end{figure}

\begin{figure}[t]
	  \centering
	  \includegraphics[width=\figwidth]{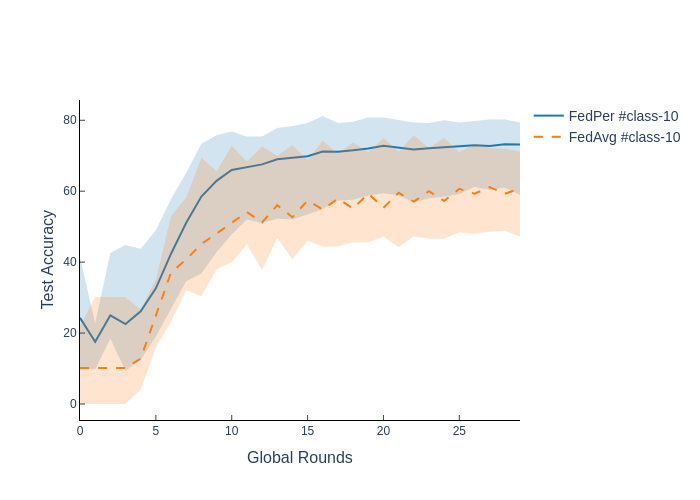}
	  \caption{Variation in the client models performance when training \mobnet[v1] on \cifar[10] using \FedAvgAlg{} and \FedPerAlg{}}
	  \label{fig:mobilenetoverlappingvariation10}
	\end{figure}

	\figuresname~\ref{fig:resnetoverlapping100}, \ref{fig:mobilenetoverlapping100} are plots on \cifar[100] corresponding to \cifar[10] plots in \sectionname~\ref{sec:exp:stat-het} and \figurename~\ref{fig:mobilenetoverlappingvariation10} is the cross-client variation in test accuracies with identical data partitioning at $k = 10$.
	
	\subsection{Effect of Personalization Layers}
\begin{figure*}
		\centering
		\begin{subfigure}[b]{0.45\textwidth}
			\centering
			\includegraphics[width=\textwidth]{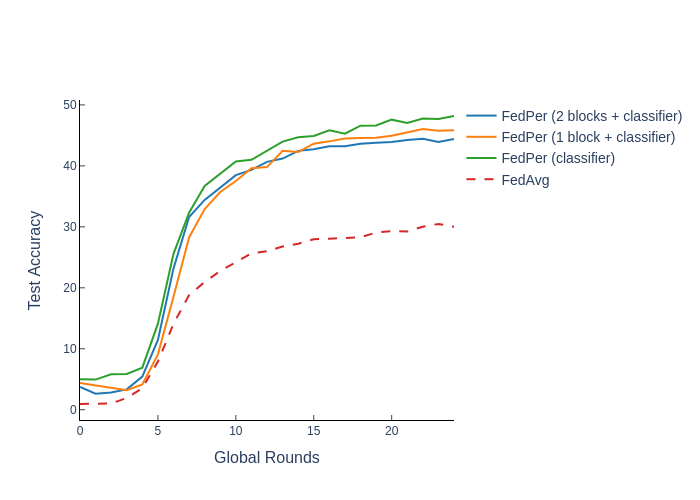}
			\caption{\mobnet[v1]}
			\label{fig:mobilenet_personal_cifar100}
		\end{subfigure}
		\hfill
		\begin{subfigure}[b]{0.45\textwidth}
			\centering
			\includegraphics[width=\textwidth]{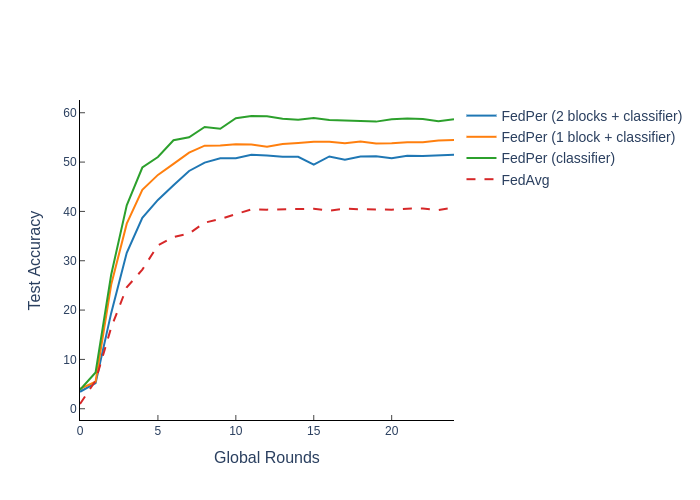}
			\caption{\resnet[34]}
			\label{fig:resnet_personal_cifar100}
		\end{subfigure}
		\caption{Performance of \FedPerAlg{} on non-identical \cifar[100] partition \wrt~\#personalization layers}
		\label{fig:personal_cifar100}
	\end{figure*}

	\figurename~\ref{fig:personal_cifar100} shows that for \cifar[100], $K_P = 1$ seems to do the best in terms of test accuracies.
	%
	
	\subsection{Effect of FineTuning}
	In the current \FedPerAlg{} approach, each global round involves the server sending the base layer parameters $\tW_B^{\bb{k-1}}$ to the clients, clients updating their local model parameters by performing e local epochs and then sending the updated base layer parameters $\tW_{B,j}^{\bb{k}}$ to the server.
	The server then aggregates the base layer parameters to get the updated $\tW_B^{\bb{k}}$.

\begin{figure}[t]
		\centering
		\includegraphics[width=\figwidth]{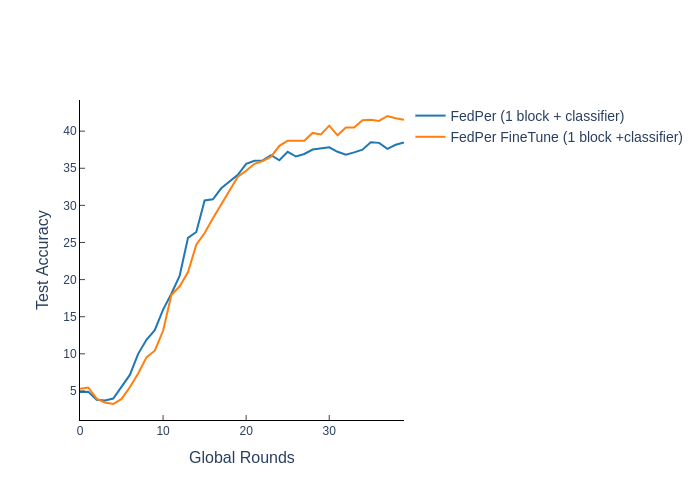}
		\caption{Effect on the performance of \mobnet[v1] on \cifar[100] with/without fine-tuning personalization layers}
		\label{fig:MobileNet_finetuning_cifar100}
	\end{figure}

\begin{figure}[t]
		\centering
		\includegraphics[width=\figwidth]{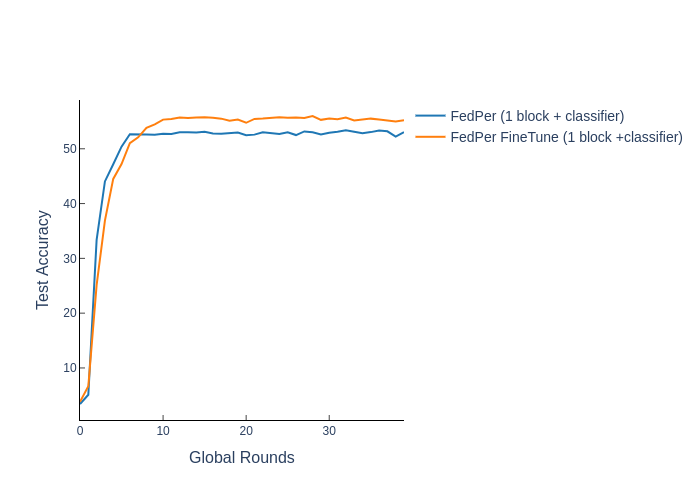}
		\caption{Effect on the performance of \resnet[34] on \cifar[100] with/without fine-tuning personalization layers}
		\label{fig:Resnet_finetuning_cifar100}
	\end{figure}

	In this experiment, at the start of each global round, we fine tune the personalization layer parameters of the clients for 1 epoch by freezing the base layers.
	We hypothesize that fine tuning the personalization layers after receiving the base layer parameters from the server will help the local client models to accommodate to the previous round changes in the base layer parameters better.
	Results of fine-tuning of \mobnet[v1] and \resnet[34] on \cifar[100] dataset are shown in \figuresname~\ref{fig:MobileNet_finetuning_cifar100} and \ref{fig:Resnet_finetuning_cifar100}.
	It can be seen from the results that fine-tuning has a positive effect on the client models by improving their accuracy thereby confirming our hypothesis.

\begin{figure}[t]
	  \centering
	  \includegraphics[width=\figwidth]{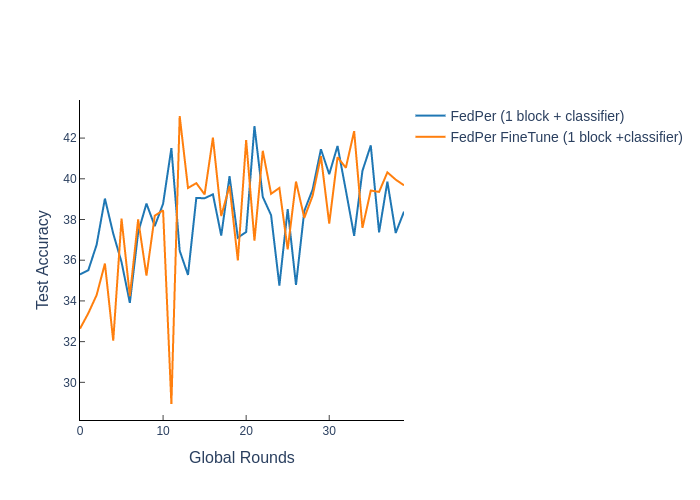}
	  \caption{Effect on the performance of \mobnet[v1] on \flickr{} with/without fine-tuning personalization layers}
	  \label{fig:MobileNet_finetuning_aes}
	\end{figure}

\begin{figure}[t]
	  \centering
	  \includegraphics[width=\figwidth]{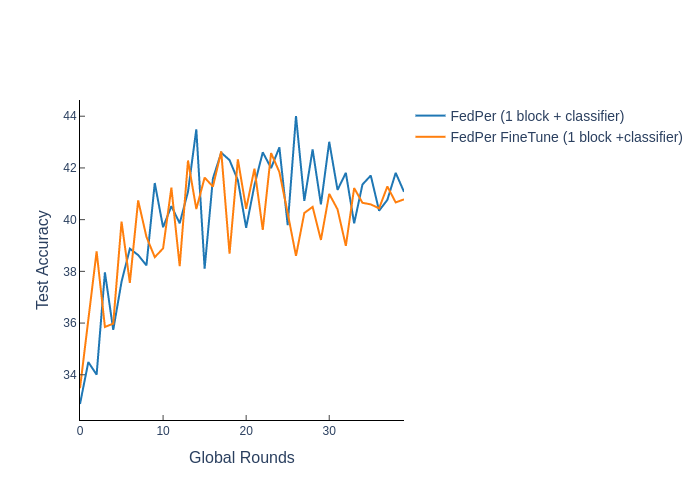}
	  \caption{Effect on the performance of \resnet[34] on \flickr{} with/without fine-tuning personalization layers}
	  \label{fig:Resnet_finetuning_aes}
	\end{figure}

	\flickr: The effect of fine-tuning on the performance of \mobnet[v1] and \resnet[34] are given in \figurename~\ref{fig:MobileNet_finetuning_aes} and \ref{fig:Resnet_finetuning_aes}.
	Interestingly fine-tuning of personalization layers does not have any noticeable effect in the performance of the client models.
	
